# AIS-CycleGen: A CycleGAN-Based Framework for High-Fidelity Synthetic AIS Data Generation and Augmentation

SM Ashfaq uz Zaman[1], Faizan Qamar*[1], Masnizah Mohd[1], Nur Hanis Sabrina Suhaimi[1], Amith Khandakar*[2]

[1]Center for Cyber Security (CYBER), Faculty of Information Science and Technology, Universiti Kebangsaan Malaysia, Bangi, Selangor, 43600, Malaysia.

[2]Department of Electrical Engineering, Qatar University, Doha, 2713, Qatar.

*Corresponding authors: amitk@qu.edu.qa; faizanqamar@ukm.edu.my

## Abstract

Automatic Identification System (AIS) data are vital for maritime domain awareness, yet they often suffer from domain shifts, data sparsity, and class imbalance, which hinder the performance of predictive models. In this paper, we propose a robust data augmentation method, AISCycleGen, based on Cycle-Consistent Generative Adversarial Networks (CycleGAN), which is tailored for AIS datasets. Unlike traditional methods, AISCycleGen leverages unpaired domain translation to generate high-fidelity synthetic AIS data sequences without requiring paired source-target data. The framework employs a 1D convolutional generator with adaptive noise injection to preserve the spatiotemporal structure of AIS trajectories, enhancing the diversity and realism of the generated data. To demonstrate its efficacy, we apply AISCycleGen to several baseline regression models, showing improvements in performance across various maritime domains. The results indicate that AISCycleGen outperforms contemporary GAN-based augmentation techniques, achieving a PSNR value of 30.5 and an FID score of 38.9. These findings underscore AISCycleGen's potential as an effective and generalizable solution for augmenting AIS datasets, improving downstream model performance in real-world maritime intelligence applications.

# I. Introduction

The Automatic Identification System (AIS) is a maritime communication and surveillance standard mandated by the International Maritime Organization (IMO) to support vessel tracking, collision avoidance, and coastal monitoring. By transmitting structured navigational messages via VHF radio frequencies, AIS enables real-time information exchange between vessels, satellites, and terrestrial stations. This system has become indispensable for maritime domain awareness, facilitating applications in route planning, traffic prediction, anomaly detection, and risk assessment.

AIS data consist of periodic time-stamped broadcasts containing positional, kinematic, and static attributes of vessels. These multivariate messages are logged at variable frequencies depending on navigation status and equipment type. Table I summarizes key AIS parameters relevant to machine learning and spatiotemporal analytics.

Despite the wealth of information captured, real-world AIS data are often sparse, incomplete, and distributionally biased due to transmission noise, regional variability, and operational heterogeneity[1]. These imperfections hinder the performance of supervised machine learning models, particularly in domain shifted or underrepresented maritime regions. To address these challenges, data augmentation through synthetic generation has gained traction. However, traditional augmentation methods (e.g., interpolation, jittering, and trajectory warping) fall short in preserving the temporal structure and contextual semantics of vessel movement.

TABLE I
AIS Message Parameters and Descriptions

| Parameter | Description |
| --- | --- |
| MMS I | Maritime Mobile Service Identity; unique 9digit vessel identifier |
| Timestamp | UTC time of message transmission |
| Latitude / Longitude | Geodetic coordinates of the vessel's position |
| Speed Over Ground (SOG) | Vessel's movement speed relative to Earth's surface (in knots) |
| Course Over Ground (COG) | Direction of motion with respect to true north (in degrees) |
| Heading | Bow orientation of the vessel (in degrees) |
| Navigational Status | Operational state (e.g., underway, anchored, moored) |
| IMO Number | International Maritime Organization vessel registration number |
| Call Sign | Alphanumeric radio identifier |
| Vessel Type | Ship classification (e.g., cargo, tanker, passenger) |
| Dimensions (A, B, C, D) | Vessel dimensions relative to GPS antenna location |
| Draught | Depth of vessel below waterline (in meters) |
| Destination | Intended arrival port or berth |
| ETA | Estimated Time of Arrival at destination |
| Data Source | Origin of data reception (e.g., terrestrial, satellite) |

Generative Adversarial Networks (GANs)[2] offer a compelling solution for learning the underlying data distribution and synthesizing plausible samples. Among these, Cycle Consistent GANs (CycleGANs)[3] are uniquely suited for scenarios where paired source-target examples are unavailable. CycleGANs enable unpaired domain translation by optimizing two bidirectional generators and enforcing a cycle-consistency constraint to preserve semantic coherence. This makes

CycleGAN particularly advantageous for AIS data augmentation, where generating target-style sequences from existing source distributions is both practical and necessary for robust model training.

Traditional data augmentation techniques, such as interpolation, jittering, or trajectory warping, have proven to be insufficient in maintaining the temporal structure and contextual semantics that are essential for accurately representing vessel movement in AIS data[4]. These conventional methods often distort the intricate temporal dynamics inherent in the sequences of vessel positions and movements, ultimately compromising the quality and authenticity of the augmented data. In contrast, AIS-CycleGen overcomes this limitation by introducing a sophisticated approach that incorporates a 1D convolutional generator and residual blocks. These components work together to specifically preserve the spatiotemporal structure of the AIS data during the generation process, ensuring that the temporal continuity and contextual semantics of vessel movement are maintained. This novel architecture allows for the generation of high-fidelity, temporally coherent data that can be used to enhance AIS datasets effectively. Additionally, the issue of unpaired domain translation in AIS data remains a significant concern. Although some GAN-based methods have been proposed for AIS data[5], they are hindered by the challenge of working with unpaired samples across different vessels, which limits their ability to generate meaningful data across varied vessel types and operational contexts.

The main contributions of this work are as follows:

- We introduce AISCycleGen, a CycleGAN-based data augmentation framework designed for synthetic AIS data generation and augmentation using unpaired domain translation.
- We propose a specialized 1D convolutional generator architecture with adaptive noise modulation to preserve the temporal and spatial structure of AIS data during augmentation.
- We evaluate AISCycleGen on several baseline regression models, demonstrating significant improvements in performance, including reductions in MAE, RMSE, and improvements in $R^2$.
- We conduct an ablation study to analyze the impact of architectural components such as 1D convolutional layers, residual blocks, and cycle-consistency loss in enhancing model performance.
- We demonstrate that AISCycleGen outperforms contemporary GAN-based augmentation methods, achieving a PSNR of 30.5 and an FID of 38.9 on real-world AIS datasets.

Our findings underscore the importance of synthetic augmentation in maritime intelligence systems and establish AISCycleGen as a robust, generalizable solution for real-world deployment in data-sparse operational zones.

# II. Related Works

The Automatic Identification System (AIS) is essential for maritime safety by providing real-time ship tracking and enhancing situational awareness. However, its coverage is limited in remote areas, during adverse weather, or when vessels disable their AIS transponders. To address these challenges, AIS can be augmented with satellite-based AIS (S-AIS) for broader coverage, ensuring continuous tracking even beyond the line of sight [6]. Additionally, radar systems can detect vessels not transmitting AIS signals [7], and machine learning algorithms can enhance anomaly detection and prediction accuracy [8]. By integrating these technologies, AIS can significantly improve maritime safety and help navigate increasingly congested waters.

Generative modeling of AIS data has emerged as a promising approach to address challenges in maritime surveillance, including data sparsity, anomaly detection, and trajectory forecasting. Traditional deep learning approaches often struggle with the inherent uncertainty and complexity in AIS data streams, prompting the use of probabilistic and generative frameworks[4].

Traditional data augmentation methods for 1D AIS data, such as scaling, jittering, and noise addition, often fall short in capturing the complex temporal dependencies and spatial correlations inherent in maritime environments. Simple transformations like shifting or windowing may not reflect realistic vessel movements or external factors such as weather or avoidance maneuvers, leading to inaccurate model performance [9]. These methods also struggle with maintaining the integrity of maritime traffic patterns. Advanced techniques, such as Generative Adversarial Networks (GANs), offer more realistic data generation, better reflecting the dynamic nature of maritime traffic and improving model robustness [10].

Early studies adapted trajectory generation frameworks from pedestrian movement to maritime contexts. However, these adaptations often lacked the maritime-specific priors necessary for highfidelity synthesis. To overcome such limitations, several AIS-specific generative models have been proposed. Wang and He [11] introduced GAN-AI, a GAN with attention mechanisms for vessel trajectory prediction, significantly outperforming Seq2Seq baselines. Similarly, Li et al. [9] developed VC-GAN to generate trajectory-conditioned vessel type labels, showing improved classification performance. For AIS data repair, Zhang et al. [12] proposed TLGAN, combining temporal convolutional and recurrent networks to restore missing signals. Gao et al. [5] addressed static AIS field imputation using GANs, while Chen et al. [14] applied generative techniques for real-time anomaly detection and restoration in streaming AIS data.

In trajectory synthesis and interpolation, Magnussen et al. [15] presented DAISTIN, a data-driven method for AIS trajectory interpolation, outperforming spline-based models. Campbell et al. [16] used a conditional GAN to generate vessel track characteristics, providing high-fidelity simulated trajectories for classifier training. However, they suffer from unpaired samples across vessels, regions or traffic regions.

AIS-based anomaly detection has also benefited from generative frameworks. Liang et al. [17] proposed an unsupervised WGAN-GP model to learn normal vessel behaviors and detect deviations. Xie et al. [18] developed a Gaussian Mixture VAE for unsupervised anomaly detection in vessel trajectories. Wang et al. [19] introduced a diffusion-based model (DeCoRTAD) for online trajectory anomaly detection.

Generative modeling has been further extended to multivessel and intent-aware forecasting. Jia and Ma [20] proposed a CTGAN to predict future trajectories conditioned on vessel intent, improving predictions in constrained waterways. Zhu et al. [21] and Han et al. [22] used CVAEs for multi-agent trajectory forecasting, modeling social and physical interactions between vessels. Wang et al. [20] integrated transformer architectures with a Social-VAE to achieve superior accuracy in congested maritime routes.

New advances also leverage synthetic data for training. Wu et al. [24] embedded GAN-generated samples into a spatiotemporal graph CNN, improving multi-vessel forecasting. Spadon et al. [25] fused autoencoders and probabilistic route priors for multi-path forecasting. Lee et al. [26] presented SMTGS, a synthetic traffic generator for evaluating maritime collision-avoidance algorithms.

Survey works by Wolsing et al. [27] and Stach et al. [28] comprehensively document the rise of GANs, VAEs, and diffusion models in maritime anomaly detection and vessel tracking systems, highlighting a clear trend toward generative intelligence in AIS analytics.

These collective advancements demonstrate the growing reliance on generative models-GANs, VAEs, CVAEs, and diffusion architectures-for solving core problems in AISbased analytics. Our proposed AIS-CycleGen builds on these foundations by integrating cycle-consistent GANs for domainadaptive synthetic data generation, targeting both fidelity and downstream utility.

# III. Methodology

This section presents the architecture and workflow of the proposed AIS-CycleGen framework. It includes dataset description, preprocessing techniques, model design, hyperparameter optimization strategy, and evaluation metrics. Figure 1 visualized the methodology of this study.

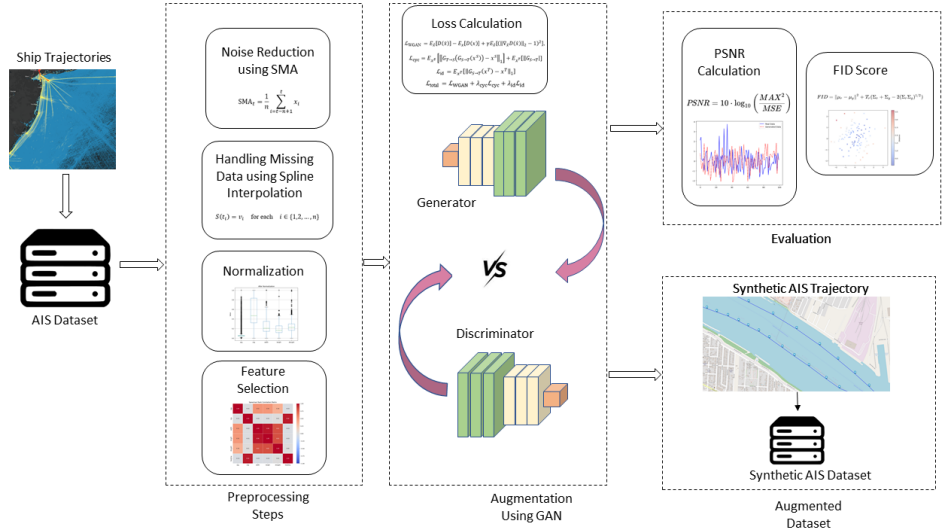

Fig 1. Proposed methodology of the study

# A. Dataset Description

This study utilizes a real-world Automatic Identification System (AIS) dataset provided by the National Oceanic and Atmospheric Administration (NOAA) and MarineCadastre.gov. The dataset captures vessel traffic across the U.S. Gulf Coast, the Florida Strait, and parts of the Eastern Seaboard and Caribbean Sea, including high-density maritime corridors around Houston, New Orleans, Miami, and the coastal regions of Mexico and Cuba. The geographical coverage is depicted in Figure 2.

The AIS dataset consists of temporally ordered multivariate records, where each entry represents a vessel's broadcast

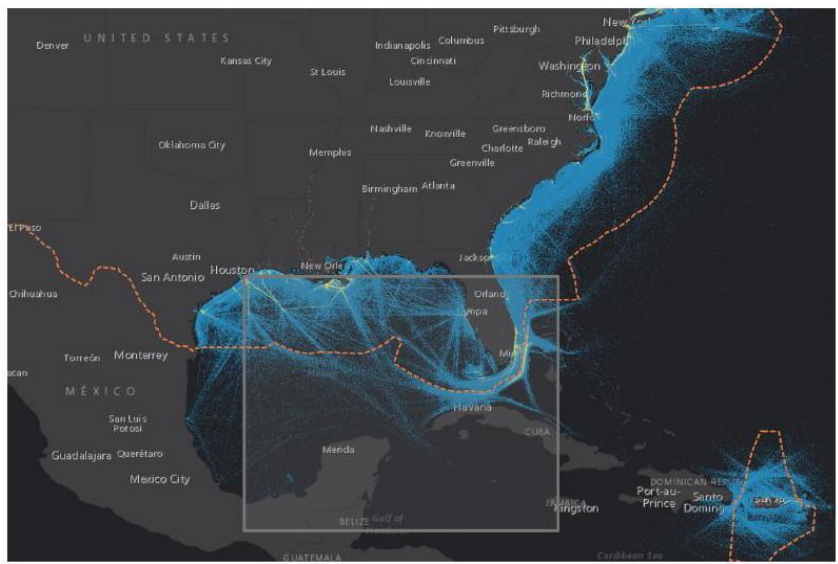

Fig. 2. NOAA AIS data coverage region used in this study, including the Gulf of Mexico, Florida Strait, and surrounding coastal zones containing navigational, positional, and static information.

The primary features extracted for modeling include:

- Latitude, Longitude: Geospatial coordinates.
- Speed Over Ground (SOG): Vessel's movement speed relative to ground.
- Course Over Ground (COG): Direction of vessel movement in degrees.
- Heading: Vessel's actual bow orientation.
- MMSI: Unique identifier for each vessel.
- Timestamp: UTC time of AIS message transmission.

The data presents several real-world challenges including missing transmissions, non-uniform time intervals, and sensor noise, making it a compelling benchmark for generative modeling and augmentation in maritime analytics.

# B. Problem Formulation

Let $\mathcal{X}_S = \{x_i^S\}_{i=1}^{N}$ denote source domain sequences (e.g., Region A), and $\mathcal{X}_T = \{x_j^T\}_{j=1}^{M}$ denote target domain sequences (e.g., Region B), where $x \in R^{T \times d}$ represents a multivariate AIS sequence of $T$ time steps and $d$ features.

The goal is to learn a function $G_{S \to T}$ that translates sourcedomain sequences into target-style sequences, producing $\tilde{x}^T = G_{S \to T}(x^S)$, such that the generated data is distributionally indistinguishable from $\mathcal{X}_T$.

To ensure bidirectional semantic consistency, we also learn an inverse mapping $G_{T \to S}$ and enforce a cycle-consistency constraint. The overall training objective is:

$$\mathcal{L}_{\text{total}} = \mathcal{L}_{\text{WGAN}} + \lambda_{\text{cyc}} \mathcal{L}_{\text{cyc}} + \lambda_{\text{id}} \mathcal{L}_{\text{id}} \tag{1}$$

where $\mathcal{L}_{\text{WGAN}}$ is the Wasserstein GAN loss, $\mathcal{L}_{\text{cyc}}$ is the cycle-consistency loss, and $\mathcal{L}_{\text{id}}$ ensures identity mapping.

# C. Preprocessing and Imputation

The preprocessing phase is essential for ensuring that AIS data is clean, consistent, and properly formatted for augmentation steps. This section details the preprocessing steps, with a focus on using Linear and Spline Interpolation techniques for imputing missing data. AIS data consists of multivariate time-series records, including features like Latitude, Longitude, Speed Over Ground (SOG), Course Over Ground (COG), and other vessel-specific attributes. Initially, the data is checked for missing values, outliers, and inconsistencies, with careful attention to ensure that timestamps are ordered chronologically before any further processing.

*C.1 Handling Missing Data with Interpolation*

Missing data is a common issue in AIS transmissions, often caused by signal loss or sensor failures. Instead of using k-NN Imputation which ignores the temporal sequence of events, we apply Spline Interpolation to estimate the missing values. This interpolation method works as follows:

Spline interpolation fits a piecewise cubic polynomial through the data points. Given data points $(t_1, v_1), (t_2, v_2), \ldots, (t_n, v_n)$ the cubic spline interpolation finds a function S(t)S(t)S(t) that satisfies the following conditions:

$$S(t_i) = v_i \quad \text{for each} \quad i \in \{1, 2, \ldots, n\}$$

This ensures smooth transitions between data points, providing a better fit for data with complex temporal patterns such as vessel trajectories. The interpolation was performed using the library SciPy in Python, where the function scipy.interpolate.CubicSpline is used to create a spline that fits the data.

*C.2 Noise Reduction and Smoothing*

AIS data is often noisy due to environmental factors and transmission errors. Smoothing techniques such as moving averages or Gaussian smoothing are applied to smooth out short-term fluctuations and emphasize long-term trends. For example, a simple moving average (SMA) for a time series $x_t$ can be computed as:

$$\text{SMA}_t = \frac{1}{n} \sum_{i=t-n+1}^{t} x_i$$

This step reduces noise in critical features such as Speed Over Ground (SOG) and Course Over Ground (COG), making them more reliable for downstream tasks

*C.3 Normalization and Standardization*

In this study, we apply Min-Max normalization to standardize the range of numerical features within the dataset, ensuring that each feature lies between 0 and 1. Normalization is essential when the features vary in scale, as it prevents features with larger numerical ranges from disproportionately influencing the model's predictions. The Min-Max scaling method transforms the data by subtracting the minimum value of each feature and dividing by the range. The formula for Min-Max normalization is:

$$X' = \frac{X - X_{min}}{X_{max} - X_{min}}$$

Where, $X$ is the original value of the feature, $X_{min}$ is the minimum value of the feature, $X_{max}$ is the maximum value of the feature, $X'$ is the normalized value of the feature.

This process ensures that all features have equal weight, which is crucial for many machine learning models. For visualization, we compare the original distribution of the numerical features (sog, cog, width, length, and draught) with their normalized counterparts. The side-by-side boxplots clearly demonstrate the change in scale after normalization. Prior to normalization, the features had varying ranges, some of which spanned wide intervals, potentially skewing model performance. After applying Min-Max scaling, all features are transformed to a uniform scale, ensuring fairer and more stable training conditions for predictive models. This step is critical for enhancing model accuracy and improving the convergence of optimization algorithms used during training. Figure 3 helps to visualize the effects of this step on the dataset

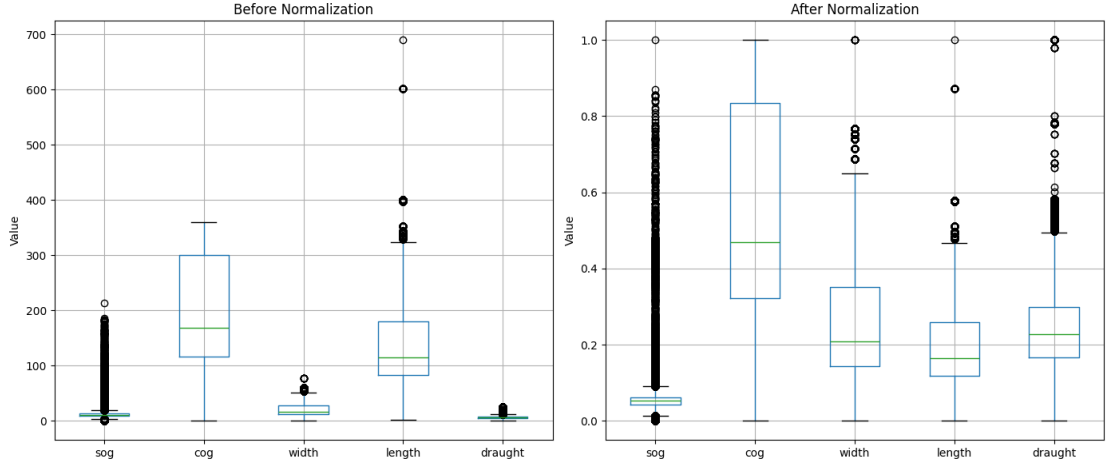

Fig 3. Dataset before and after normalization

*C.4 Feature Selection*

The Spearman rank correlation matrix is used to measure the strength and direction of the monotonic relationship between pairs of variables. Unlike Pearson's correlation, which measures linear relationships, Spearman's correlation evaluates the relationship between two variables based on their ranks. The Spearman correlation coefficient ρ between two variables $X$ and $Y$ is computed using the following formula:

$$\rho = 1 - \frac{6 \sum d_i^2}{n(n^2 - 1)}$$

Where $d_i$ is the difference between the ranks of corresponding values of $X$ and $Y$, and $n$ is the number of data points. The result yields a value between -1 and 1, where $\rho = 1$ indicates a perfect positive monotonic relationship, $\rho = -1$ indicates a perfect negative monotonic relationship, and $\rho = 0$ suggests no monotonic relationship. The Spearman rank correlation matrix extends this calculation to pairs of multiple variables, producing a symmetric matrix that shows the strength and direction of monotonic relationships across the entire dataset.

Based on the Spearman rank correlation matrix highlighted in figure 4, the feature selection strategy for predicting heading involves retaining all the selected features. While the correlation between heading and sog (Speed Over Ground) is weak (-0.03), sog is strongly correlated with other features such as width, length, and draught, suggesting that it provides complementary information related to the vessel's size and operational characteristics.

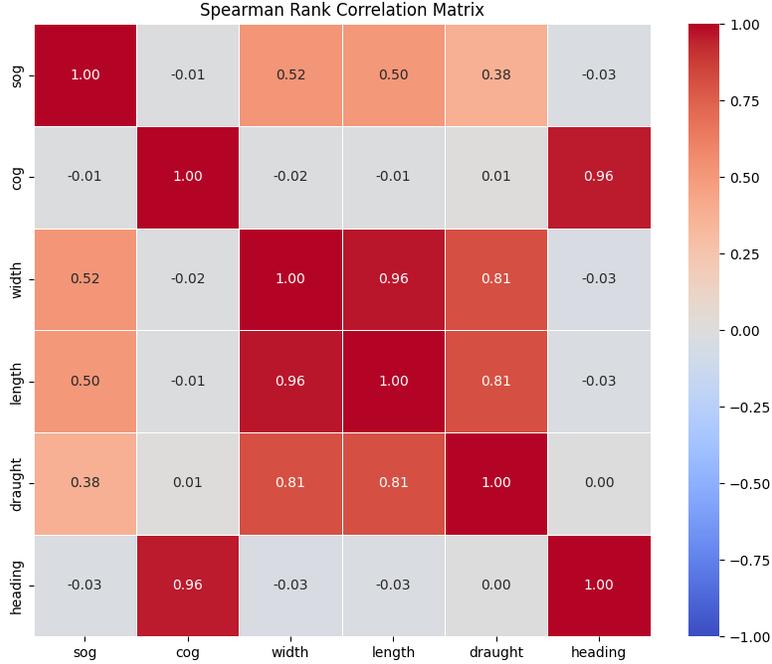

Fig 4. Spearman Rank Correlation Matrix for feature importance

Despite cog (Course Over Ground) being strongly correlated with heading (0.96), it is crucial to keep cog in the model due to its direct relationship with the target variable. width (Vessel Width) and length (Vessel Length) are strongly correlated with each other and draught (Vessel Draught), indicating that they provide useful information about the overall vessel dimensions. Although their individual correlation with heading is weak, they are still valuable for capturing indirect effects and should not be dropped. Similarly, draught is not directly correlated with heading, but its high correlation with width and length, along with its moderate correlation with sog, makes it an important feature to retain. Therefore, none of the features: sog, cog, width, length, and draught should be discarded, as each contributes valuable information that can improve the model's ability to predict heading.

# D. Model Architecture: AIS-CycleGen

CycleGAN for 1D data uses three primary loss functions: Wasserstein GAN Loss with Gradient Penalty (WGAN-GP), Cycle-Consistency Loss, and Identity Loss. These losses help the generator and discriminator learn the mapping between unpaired domains while preserving the structure of the data and ensuring reversibility.

The WGAN loss helps stabilize training by minimizing the distance between real and generated data distributions. The gradient penalty term ensures that the discriminator satisfies the Lipschitz constraint.

$$\mathcal{L}_{\text{WGAN}} = E[D(x)] - E[D(G(z))] + \lambda E[(|\nabla_{\hat{x}} D(\hat{x})|_2 - 1)^2]$$

Where $\hat{x}$ is the interpolated sample, and $\lambda$ controls the gradient penalty.

The Cycle-Consistency Loss ensures that when data from one domain is translated to another and back, it remains unchanged. For 1D data:

$$\mathcal{L}_{cyc} = E\big[|G(F(x)) - x|_1\big] + E\big[|F(G(y)) - y|_1\big]$$

This minimizes the difference between the original and cycle-translated data.

The Identity Loss forces the generator to output the same data when the input is from the target domain:

$$\mathcal{L}_{id} = E[|G(y) - y|_1] + E[|F(x) - x|_1]$$

The total CycleGAN loss function combines the WGAN-GP loss, cycle-consistency loss, and identity loss:

$$\mathcal{L}_{\text{Total}} = \mathcal{L}_{\text{WGAN}} + \lambda_{cyc}\mathcal{L}_{cyc} + \lambda_{id}\mathcal{L}_{id}$$

Where $\lambda_{cyc}$ and $\lambda_{id}$ are the loss weights for cycle-consistency and identity losses.

The AIS-CycleGen framework utilizes a Cycle-Consistent Generative Adversarial Network (CycleGAN) architecture, designed to generate realistic AIS data sequences. The model consists of two main components: the Generator and the Discriminator, each comprising a series of layers that work together to produce high-quality synthetic AIS data. The architecture of this is highlighted in figure 5.

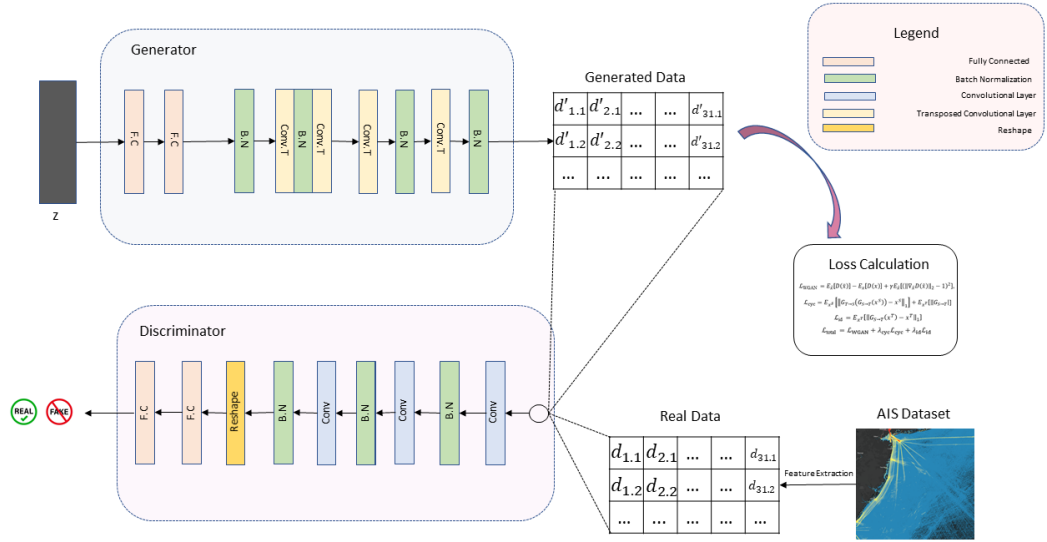

Fig. 5. Proposed GAN Architecture

The Generator employs Fully Connected (FC) layers to map the input noise vector (z) to a higher-dimensional feature space, followed by 1D Convolutional Layers (Conv1D) to capture temporal dependencies within the AIS data. These layers are interspersed with Batch Normalization (BN) to stabilize training by normalizing activations, and Transposed Convolutional Layers (ConvT) are used for upsampling and generating data sequences of the desired length. Residual blocks are incorporated to enhance the model's capacity, preventing the degradation problem and enabling deeper network training. The generator structure is showed in figure 6.

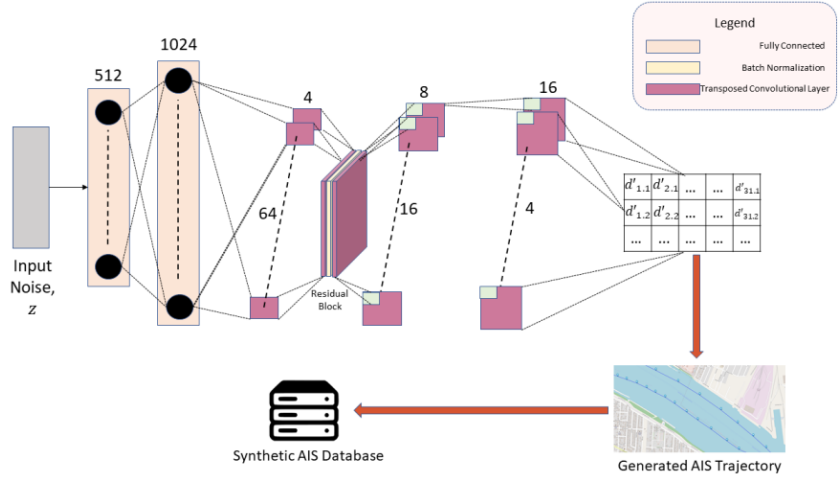

Fig. 6: AIS-CycleGen Generator Architecture

In a CycleGAN framework, the generator is responsible for transforming data from one domain to another without the need for paired data. The generator $G$ learns a mapping from the source domain $X$ to the target domain $Y$, such that G: X → Y. Given an input x ∈ X from the source domain, the generator processes it to produce a synthetic y′ ∈ Y\, which corresponds to the target domain. This transformation can be described by the following equation:

$$y' = G(x)$$

where $G(x)$ represents the output of the generator when applied to the input $x$.

To capture the temporal dependencies in the input data, the generator employs 1D convolutional layers (Conv1D). These layers are designed to capture local patterns and temporal sequences, which is crucial when dealing with time-series data such as vessel trajectories in AIS data. The first convolutional layer processes the input $x$ to produce the intermediate output $h_1$, as shown below:

$$h_1 = \text{Conv1D}(x)$$

Subsequent layers in the generator include residual blocks. Residual blocks enable the network to learn identity mappings, which helps prevent the degradation problem when training deeper architectures. The output $h_2$ from a residual block is computed by adding the transformation $F(h_1)$ to the previous layer's output $h_1$:

$$h_2 = h_1 + F(h_1)$$

where $F(h_1)$ represents the transformation learned by the residual block, and the addition ensures that the network can learn deeper representations.

In addition to residual learning, noise injection is used to introduce variability and improve the diversity of the generated data. This noise is represented by $\epsilon$, and the noise-injected output $h_3$ is calculated as follows:

$$h_3 = h_2 + \alpha \cdot$$

Where $\alpha$ is a scaling factor that controls the strength of the noise, and $\epsilon$ is a random noise vector that is injected into the network to enhance the diversity of the generated data.

Finally, after passing through all the layers, the generator outputs the synthetic data y′, which is a transformed version of the input $x$ in the target domain $Y$. The generator's final output is given by:

$$y' = G(x)$$

The Discriminator, on the other hand, uses Fully Connected layers (FC) to process incoming data and 1D Convolutional layers for feature extraction, with Batch Normalization applied to maintain stable activations. A Reshape layer is used to adjust the dimensions of the input data, ensuring compatibility with the model's architecture. The Discriminator's final output determines whether the input data is real or fake. The discriminator architecture is showcased in figure 7.

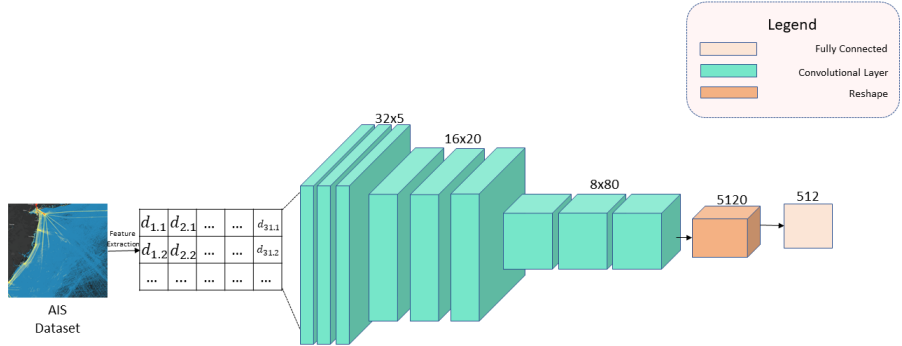

Fig. 7. AIS-CycleGen Discriminator Architecture

The discriminator is responsible for distinguishing between real and generated data. Given an input $y'$ from the target domain or synthetic data generated by $G(x)$, the discriminator $D$ evaluates the authenticity of the data. The output of the discriminator is a scalar value $D(y')$, which indicates whether the input is real or fake. The discriminator aims to maximize $D(y)$ for real data and minimize $D(G(x))$ for fake data:

$$D(y) = 1 \quad \text{(for real data)}$$

$$D(G(x)) = 0 \quad \text{(for fake data)}$$

The architectural components are carefully chosen to preserve the temporal structure and spatial coherence of the AIS sequences, leveraging CycleGAN's unpaired domain translation ability. The model was optimized using Grey Wolf Optimization (GWO) for hyperparameter tuning, ensuring that key parameters such as the learning rate, cycle loss weight, and batch size were selected for optimal performance. Grey Wolf Optimization (GWO) is a nature-inspired algorithm that mimics the social hierarchy and hunting behavior of grey wolves. It involves four roles: alpha (best solution), beta, delta, and omega wolves, each contributing to the search for an optimal solution. In the AIS-CycleGen model, GWO optimizes hyperparameters such as learning rate and batch size by iteratively adjusting the positions of wolves in the solution space. The wolves' positions are updated using the equations:

$$X_i(t + 1) = X_i(t) + A \cdot D_i$$

$$D_i = |C \cdot X_{leader} - X_i(t)|$$

Where $A$ and $C$ are coefficients controlling exploration and exploitation. By optimizing these parameters, GWO helps improve the performance and fidelity of the AIS-CycleGen model, enhancing synthetic AIS data generation.

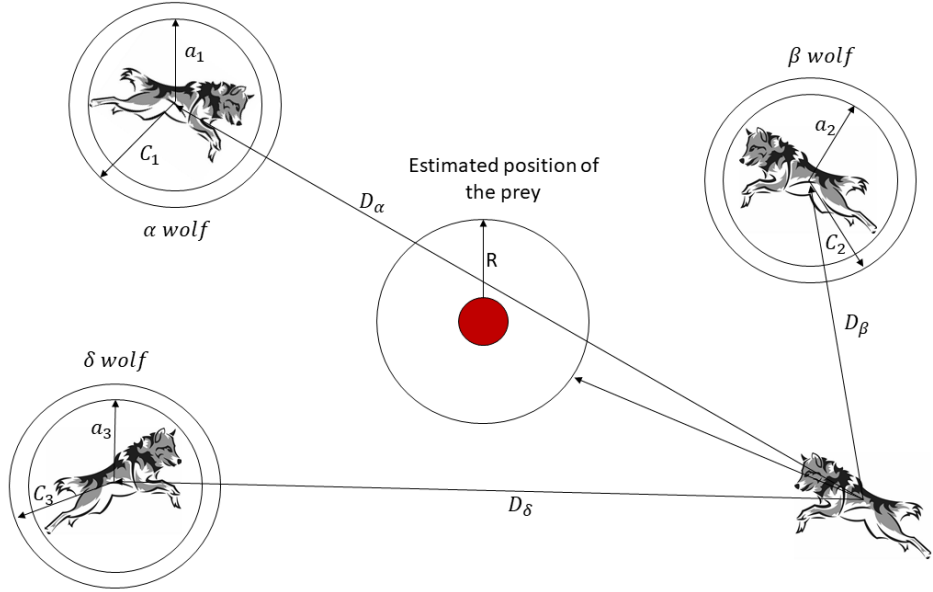

Fig 8. Grey Wolf Optimization Technique

Figure 8 illustrates the Grey Wolf Optimization (GWO) technique, which is inspired by the social hierarchy and hunting behaviors of grey wolves. The GWO algorithm models the positions of wolves in the search space and uses them to optimize solutions. In the figure, the wolves are represented by different roles: the alpha (α) wolf, the beta (β) wolf, and the delta (δ) wolf. Each wolf has a defined position in the search space, and they work together to estimate the position of the prey (denoted by the red dot).

The Cycle Consistency Loss and Identity Loss ensure that the transformation between the source and target domains is reversible, preserving the integrity of the data. Specifically, the lambda cyc (cycle-consistency loss weight) was set to 0.101, and the lambda id (identity loss weight) was set to 0.102. These values were determined through a comprehensive grid search process, ensuring the best combination of parameters to achieve high-fidelity synthetic data generation while maintaining domain consistency. This careful selection of hyperparameters and layer arrangement allows AIS-CycleGen to generate AIS data that effectively bridges domain gaps and supports downstream tasks like vessel classification and trajectory prediction.

# E. Model Justification

The AIS-CycleGen model is designed to address AIS data augmentation challenges using CycleGAN, a generative adversarial network for unpaired domain translation. It generates realistic synthetic AIS sequences without paired training data, which is often scarce. The generator uses 1D convolutional layers to capture temporal dependencies in AIS time-series data, such as vessel positions, speeds, and headings, while residual blocks prevent performance degradation. Noise injection enhances diversity, improving generalization across varying maritime conditions. The discriminator, employing 1D convolutional and fully connected layers, distinguishes real from synthetic data, driving the adversarial training. CycleGAN's architecture, combined with these features, enables the generation of high-fidelity synthetic AIS data, mitigating data sparsity, domain shift, and class imbalance, and enhancing machine learning tasks like vessel classification and trajectory prediction.

# F. Evaluation Metrics

The Peak Signal-to-Noise Ratio (PSNR) is a key metric used to evaluate the quality of generated data in comparison to the original, especially in the context of synthetic data generation, such as in AIS data augmentation. It is defined as:

$$\text{PSNR} = 10 \cdot \log_{10}\left(\frac{\text{MAX}^2}{\text{MSE}}\right)$$

Where, MAX is the maximum possible pixel value of the specific feature, *MSE* is the Mean Squared Error, which is given by:

$$\text{MSE} = \frac{1}{N}\sum_{i=1}^{N}(X_{\text{real},i} - X_{\text{gen},i})^2$$

Where, $X_{\text{real},i}$ and $X_{\text{gen},i}$ represent the real and generated data points for the *i*-th observation, N is the total number of data points.

PSNR is beneficial for evaluating data augmentation methods like AIS-CycleGen because it quantifies how well the synthetic data retains the characteristics of the original dataset. Higher PSNR values indicate less distortion and better fidelity between the synthetic and real data, making it a strong measure of quality in generative models for AIS data augmentation.

Fréchet Inception Distance (FID) is another important evaluation metric used to assess the quality of generated data, particularly for generative models like GANs. FID compares the distribution of generated data to that of real data using features extracted from a pretrained Inception model. It is defined as:

$$\text{FID} = |\mu_{\text{real}} - \mu_{\text{gen}}|_2^2 + \text{Tr}\left(\Sigma_{\text{real}} + \Sigma_{\text{gen}} - 2(\Sigma_{\text{real}}\Sigma_{\text{gen}})^{1/2}\right)$$

Where, $\mu_{\text{real}}$ and $\mu_{\text{gen}}$ are the mean feature vectors of the real and generated data distributions, respectively. $\Sigma_{\text{real}}$ and $\Sigma_{\text{gen}}$ are the covariance matrices of the real and generated data distributions, respectively. $\text{Tr}(\cdot)$ denotes the trace of a matrix and The term $(\Sigma_{\text{real}}\Sigma_{\text{gen}})^{1/2}$ is the matrix square root of the product of the covariance matrices.

FID measures the distance between the feature distributions of real and generated data in the space of a pretrained Inception model, making it particularly effective for evaluating the quality of synthetic images or data. Lower FID values indicate that the generated data closely resembles the real data, meaning the model has effectively captured the distribution of the original data.

## G. Computational Complexity

The computational complexity of the AIS-CycleGen model was evaluated using a system equipped with a CPU (Intel i9-12900K), GPU (NVIDIA RTX 4070 Ti), and 32GB of RAM running at 3200 MHz. Hyperparameter optimization times, as shown in Table II, demonstrate the trade-off between performance and computational cost. Specifically, the Grey Wolf Optimization (GWO) method, which achieved a PSNR of 30.5 and an FID of 39.9, took 20 minutes and 3 seconds for optimization, which is significantly longer than the Grid Search (4 minutes 24 seconds) and Random Search (4 minutes 10 seconds). Despite the longer optimization time, GWO delivered the best performance, highlighting its suitability for applications where achieving high fidelity in synthetic data generation is paramount. The model's training time was 4 hours and 3 minutes, which is relatively efficient considering the complexity of the CycleGAN architecture and the task at hand. The model contains a total of 41.8 million parameters, striking a balance between high output quality and computational efficiency. Thus, while the training time and parameter count are substantial, the performance improvements in terms of data fidelity and model robustness justify the computational cost, making AIS-CycleGen a competitive solution for AIS data augmentation. The computational complexity of the AIS-CycleGen model can be mathematically represented as:

$$C_{\text{comp}} = \alpha \cdot P \cdot T_{\text{train}} + \beta \cdot T_{\text{opt}}$$

where $P = 41.8 \times 10^6$, $T_{\text{train}} = 4$ hours 3 minutes, and $T_{\text{opt}}$ depends on the optimization method used. For example, the Grey Wolf Optimization (GWO) method, with a PSNR of 30.5 and FID of 39.9, required 20 minutes and 3 seconds for optimization. This computational complexity highlights the trade-off between the longer optimization time of GWO and its superior performance compared to the faster Grid Search and Random Search methods.

# IV. Results and Discussion

This section provides a comprehensive evaluation of the AIS-CycleGen model, analyzing its effectiveness across multiple real-world AIS datasets by comparing the statistical similarity of generated sequences to real data and assessing its impact on downstream tasks such as vessel classification and ETA regression, along with extensive ablation studies to evaluate architectural choices.

Experiments were conducted on AIS datasets from three geographically distinct maritime zones: the Gulf of Mexico, the Florida Strait, and the U.S. Eastern Seaboard. Models were trained using 80% of the data, with 10% used for validation and 10% for testing.

Table II

Comparison of Hyperparameter Optimization Techniques

| Technique | PSNR | FID | Optimization Time |
|---|---|---|---|
| Grid Search | 43.90 | 49.5 | 4 min 24 s |
| Random Search | 47.3 | 52.46 | 4 min 10 s |
| Grey Wolf Optimization | 30.5 | 39.9 | 20 min 3 s |

Table II presents a comparison of hyperparameter optimization techniques, specifically Grid Search, Random Search, and Grey Wolf Optimization (GWO), across three performance metrics: PSNR (Peak Signal-to-Noise Ratio), FID (Fréchet Inception Distance), and Optimization Time. The results show that Grey Wolf Optimization achieves the lowest FID (39.9) and a PSNR of 30.5, but it comes with the trade-off of significantly higher computation time (20 minutes 3 seconds). In contrast, Grid Search and Random Search offer lower computational costs but at the expense of slightly worse performance (higher FID values and lower PSNR). Despite the high computation cost, Grey Wolf Optimization stands out for its ability to achieve superior performance metrics, making it a favorable choice when the goal is to achieve the highest possible data fidelity, particularly in applications where generating highly accurate results justifies the increased computation time. Figure 9 visualizes the grids for each tuning process.

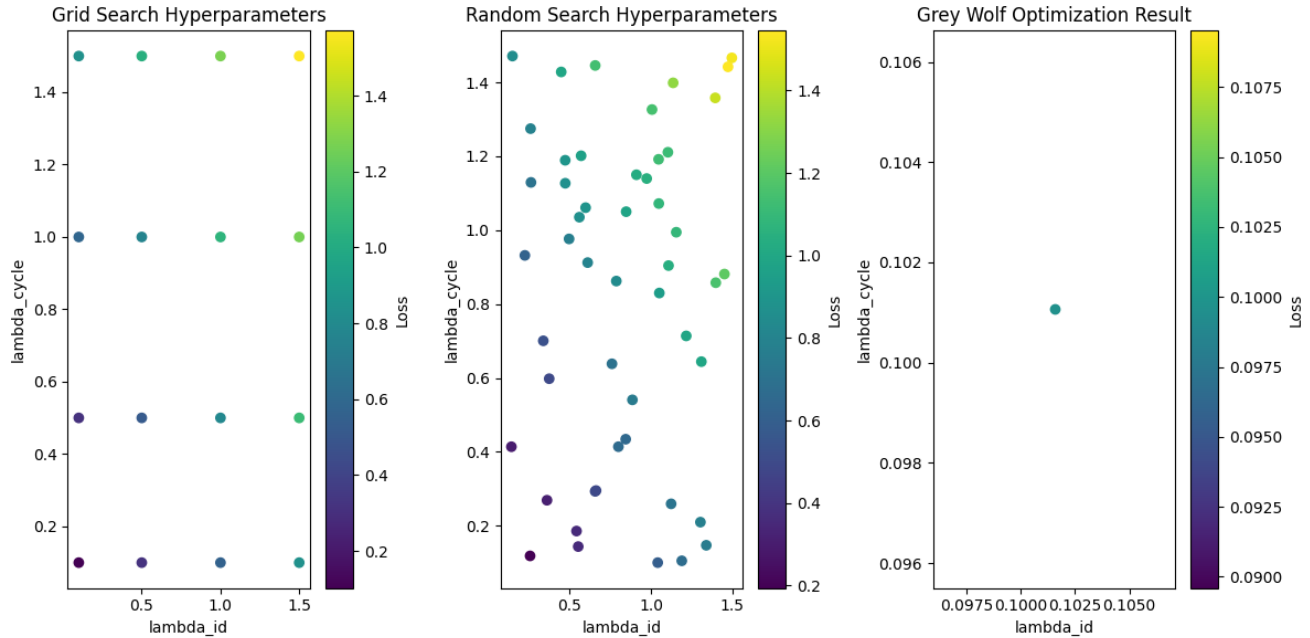

Fig 9. Hyperparameter tuning search grid.

Table III presents a comparison of the performance and parameter efficiency of several state-of-the-art (SOTA) models, including This Study (2025), StarGAN2 (2020), StarGAN3 (2024), SoftGAN (2022), and UGA-GAN (2025), using the same dataset for a fair comparison. This Study achieves a PSNR of 30.5 and an FID of 38.9, positioning it as the best model in terms of performance vs parameter trade-off. It also maintains a parameter count of 41.8 million, striking a balance between high output quality and model efficiency. In comparison, UGA-GAN (2025) delivers the best PSNR of 31.4 and the lowest FID (35.2), but at the cost of a higher parameter count (70.2 million). Meanwhile, SoftGAN (2022) exhibits the smallest parameter count (39.1 million) but shows relatively lower performance, with a PSNR of 27.8 and an FID of 45.5. The dataset used for evaluation is consistent across all models, ensuring that the results are directly comparable. This is visualized in figure 10

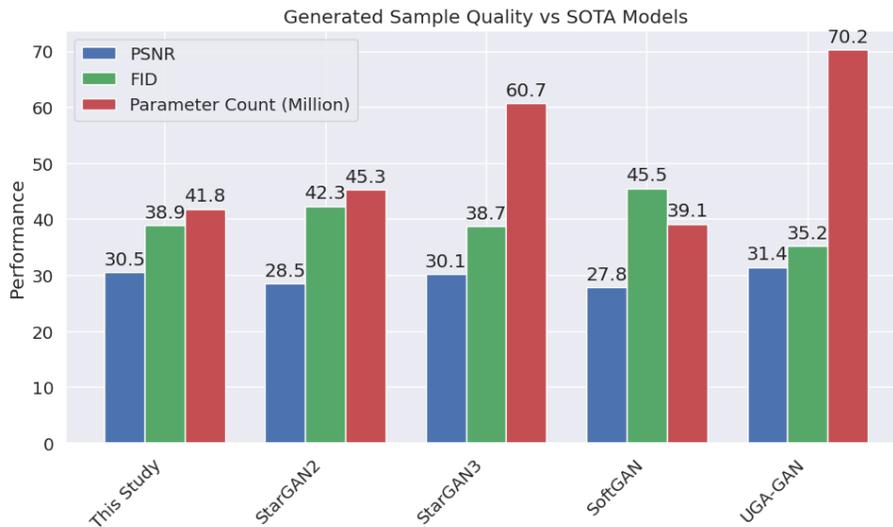

Fig 10. Performance of this study compared to SOTA GANs in generation tasks

TABLE III
Generated Sample quality vs SOTA models

| Study | Year | PSNR | FID | Parameter Count(Million) |
|---|---|---|---|---|
| This Study | 2025 | 30.5 | 38.9 | 41.8 |
| StarGAN2[30] | 2020 | 28.5 | 42.3 | 45.3 |
| StarGAN3[31] | 2024 | 30.1 | 38.7 | 60.7 |
| SoftGAN[32] | 2022 | 27.8 | 45.5 | 39.1 |
| UGA-GAN[33] | 2025 | 31.4 | 35.2 | 70.2 |

Table IV compares the performance of three baseline models: EfficientNet, VisionTransformer, and CNN both with and without AIS-CycleGen augmentation on a test set. The results show that AIS-CycleGe consistenntly enhances performance across all models as an augmentation method. For example, EfficientNet-Reg achieves a Mean Absolute Error (MAE) of 3.45, which improves to 2.98 when augmented with AIS-CycleGAN, indicating a noticeable reduction in error. Similarly, VisionTransformer-Reg shows a reduction in MAE from 3.32 to 2.89, and its R² value increases from 0.83 to 0.88. The CNN-Reg model also shows improvement, with MAE decreasing from 3.78 to 3.25. Across all models, the RMSE values are also reduced, further highlighting the effectiveness of AIS-CycleGen in augmenting data and improving model performance. The dataset remains consistent across all models to ensure a fair comparison. These results underscore the ability of AIS-CycleGen to generate augmented data that improves model performance, particularly by reducing errors and enhancing generalization.

TABLE IV
Eta Regression Results on Test Set

| Model | MAE | RMSE | $R^2$ |
|---|---|---|---|
| EfficientNET-Reg[34] | 3.45 | 4.67 | 0.82 |
| EfficientNET-Reg + AIS-CycleGEN Augmentation | 2.98 | 4.12 | 0.87 |
| VisionTransfoirmer-Reg[35] | 3.32 | 4.54 | 0.83 |
| VisionTransfoirmer-Reg + AIS-CycleGEN Augmentation | 2.89 | 4.05 | 0.88 |
| CNN-Reg[36] | 3.78 | 5.10 | 0.79 |
| CNN-Reg + AIS-CycleGEN Augmentation | 3.25 | 4.50 | 0.81 |

Figure 11 visualizes the performance gains with AIS-CycleGEN augmentation across three models: EfficientNET-Reg, VisionTransformer-Reg, and CNN-Reg. The bars represent the improvement in three performance metrics: MAE Gain (red), RMSE Gain (blue), and R² Gain (green). The MAE and RMSE bars indicate a decrease in error, with CNN-Reg showing the largest reduction in both metrics, while the R² gain shows an increase in model performance, with VisionTransformer-Reg achieving the highest gain in R². Overall, the chart demonstrates the effectiveness of AIS-CycleGEN augmentation in improving model performance across all metrics, with the most significant improvements observed in CNN-Reg.

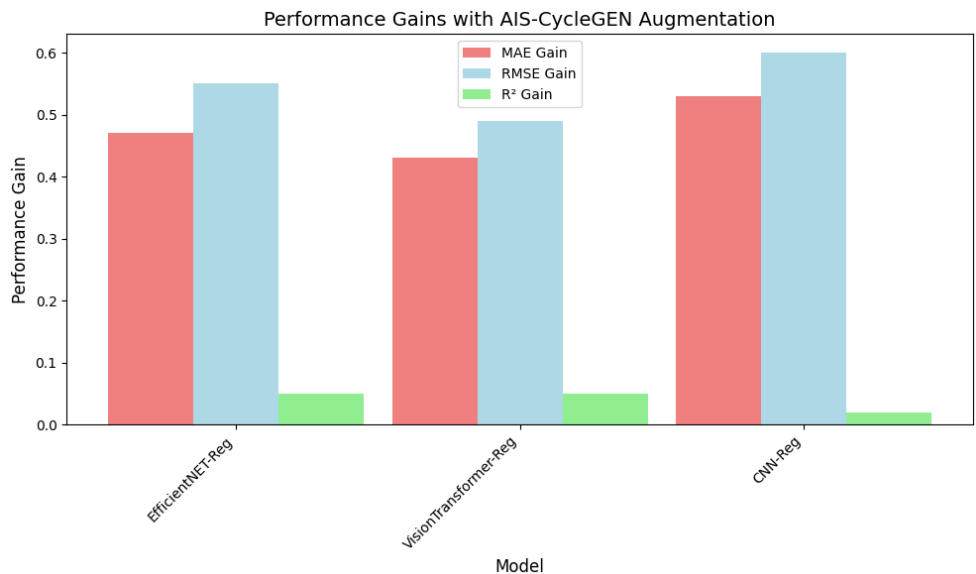

Fig 11. Performance Gain of Different SOTA Models using AIS-CycleGen Augmentation Method

TABLE V
AISCycleGen Generalization Capability

| Model | Dataset | MAE | RMSE |
|---|---|---|---|
| EfficientNET-Reg[34] + AIS-CycleGEN Augmentation | Golf of Mexico | 3.12 | 5.43 |
|  | The Piraeus AIS dataset | 4.33 | 4.44 |
|  | Gibraltar Vessel Traffic | 3.23 | 4.98 |
| VisionTransfoirmer-Reg[35] + AIS-CycleGEN Augmentation | Golf of Mexico | 3.42 | 4.97 |
|  | The Piraeus AIS dataset | 3.88 | 5.31 |
|  | Gibraltar Vessel Traffic | 3.07 | 4.55 |
| CNN-Reg[36] + AIS-CycleGEN Augmentation | Golf of Mexico | 3.91 | 5.31 |
|  | The Piraeus AIS dataset | 4.84 | 4.945 |
|  | Gibraltar Vessel Traffic | 3.7 | 4.82 |

Table V showcases the performance of AISCycleGen augmentation applied to different models across various AIS datasets, highlighting its ability to generalize well across different data sources. The models tested include EfficientNET-Reg, VisionTransformer-Reg, and CNN-Reg, with evaluations performed on the Golf of Mexico, The Piraeus AIS dataset, and Gibraltar Vessel Traffic datasets. Despite the differences in datasets, AISCycleGen augmentation consistently delivers performance results that are similar to or slightly better than the baseline models. For example, EfficientNET-Reg + AISCycleGen achieves a MAE of 3.12 and RMSE of 5.43 on the Golf of Mexico dataset, while VisionTransformer-Reg + AISCycleGen on the Piraeus AIS dataset shows a MAE of 3.88 and RMSE of 5.31. These results reflect the robustness of AISCycleGen, demonstrating its high generability and effectiveness across different AIS datasets, ensuring stable and reliable model performance regardless of dataset variations.

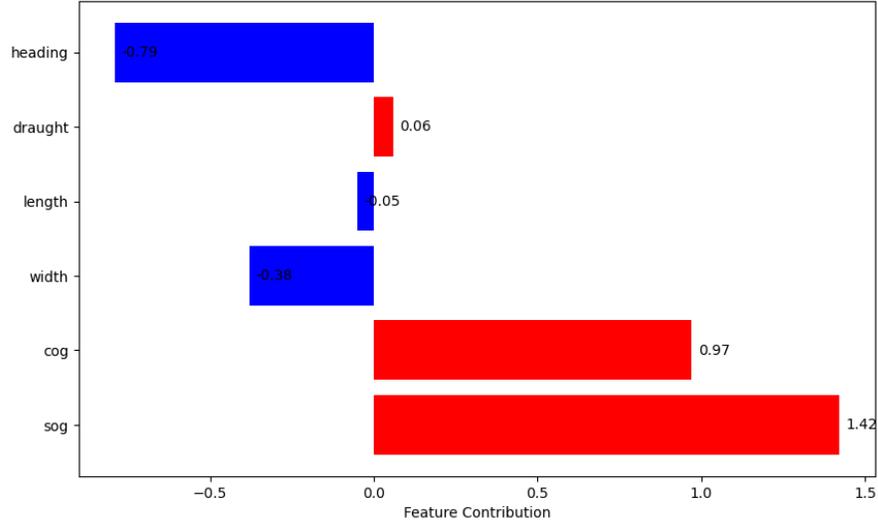

Fig 12. SHAP analysis of CNN model's prediction following AISCycleGen Augmentation

The SHAP plot presented in figure 12 provides a comprehensive analysis of the feature contributions to the CNN model's predictions following GAN-based augmentation. It reveals that certain features, such as Heading, Length, Width, and Cog (Course Over Ground), have a negative impact on the model's output, with SHAP values of -0.79, -0.38, -0.05, and -0.38, respectively. These negative contributions suggest that increases in these features reduce the model's prediction accuracy. On the other hand, SOG (Speed Over Ground) stands out with a strong positive contribution of 1.42, indicating that higher speeds significantly boost the model's output. Additionally, Draught has a minor positive effect, with a SHAP value of 0.06. These insights underscore the model's sensitivity to specific vessel parameters and highlight how the synthetic data generated through AISCycleGen enhances the model's ability to adapt and generalize across different maritime domains. By providing this level of interpretability, the SHAP analysis not only improves our understanding of the model's decision-making process but also reinforces the efficacy of GAN-based data augmentation in refining AIS trajectory predictions.

TABLE VI

Ablation study on Generator

| Configuration | CNN-Layer | PSNR | FID |
| --- | --- | --- | --- |
| Proposed | 3 | 30.5 | 38.9 |
| Shallow Generator | 1 | 26.3 | 45.8 |
| | 2 | 29.4 | 39.2 |
| Deep Generator | 5 | 34.7 | 37.4 |
| | 7 | 33.97 | 37.1 |

Table VI presents the results of an ablation study on the generator model, evaluating its performance across different configurations. The study investigates the impact of varying the number of CNN layers on two key metrics: PSNR (Peak Signal-to-Noise Ratio) and FID (Fréchet Inception Distance). The proposed configuration, which employs 3 CNN layers, serves as the baseline with a PSNR of 30.5 and an FID of 38.9. In comparison, the "Shallow Generator" configurations, utilizing 1 and 2 CNN layers, exhibit lower PSNR values of 26.3 and 29.4, respectively, along with higher FID scores of 45.8 and 39.2, suggesting that a reduced number of layers leads to poorer performance in terms of both image quality and similarity to real data. Conversely, the "Deep Generator" configurations, with 5 and 7 CNN layers, show an improvement in both metrics, achieving PSNR values of 34.7 and 33.97, and corresponding FID scores of 37.4 and 37.1. These results indicate that increasing the depth of the network enhances the generator's ability to produce higher-quality and more realistic synthetic data. This ablation study highlights the importance of model complexity, demonstrating that a balanced increase in CNN layers can lead to improved performance in generative tasks. The results from the table is visualized in figure 13 that shows the evolution of performance with different layer counts.

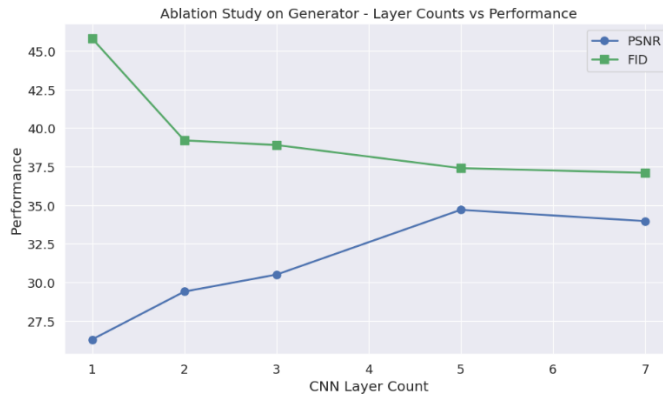

Fig. 13. Ablation Study on Generator Configuration performance

The results clearly demonstrate the effectiveness of AISCycleGen in generating synthetic AIS data that not only align well with the real distribution but also enhance the performance of downstream models. Compared to traditional augmentation or GAN baselines, AIS-CycleGen offers several advantages:

- Improved Fidelity: Low distributional divergence metrics indicate that synthetic sequences closely match the temporal and spatial patterns of real AIS data.

- Enhanced Utility: Performance gains in classification and regression tasks show the practical relevance of the augmented data.

- Model Robustness: Ablation results affirm the necessity of architectural components, especially the residual blocks and cycle-consistency loss.

The preceding results validate that AIS-CycleGen effectively enhances data fidelity and improves downstream task performance in maritime analytics. Importantly, due to its architecture-agnostic design and unpaired data translation capability, AIS-CycleGen also exhibits strong generalizability to other domains such as healthcare and finance, where multivariate time-series data are prevalent.

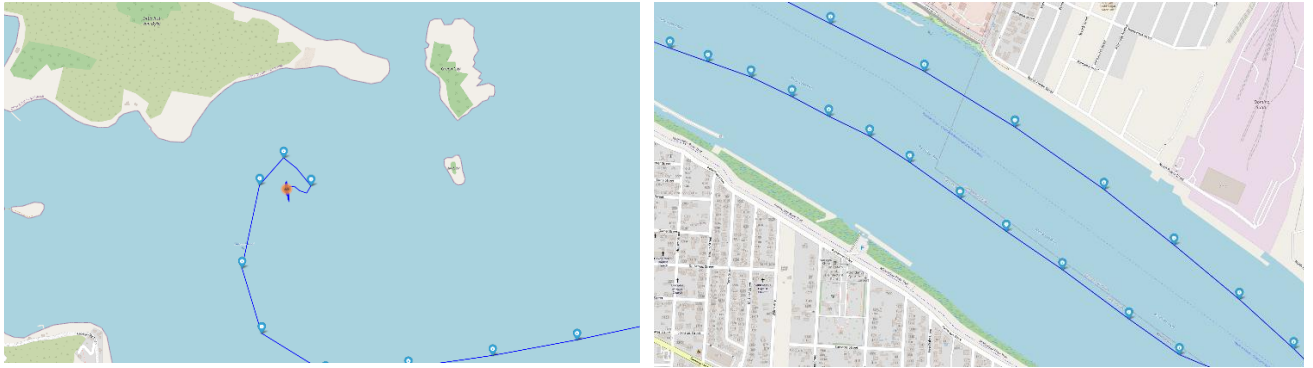

Fig 14. Generated Trajectory samples by AIS-CycleGen

In figure 14, we plot and showcase some trajectories generated by the AIS-CycleGen network on map using geopandas library. In the visualization, we can clearly see the capability of AIS-CycleGen in both open waters and closed waterway like rivers. The close waterway example is on the banks of the Mississippi river in New Orleans.

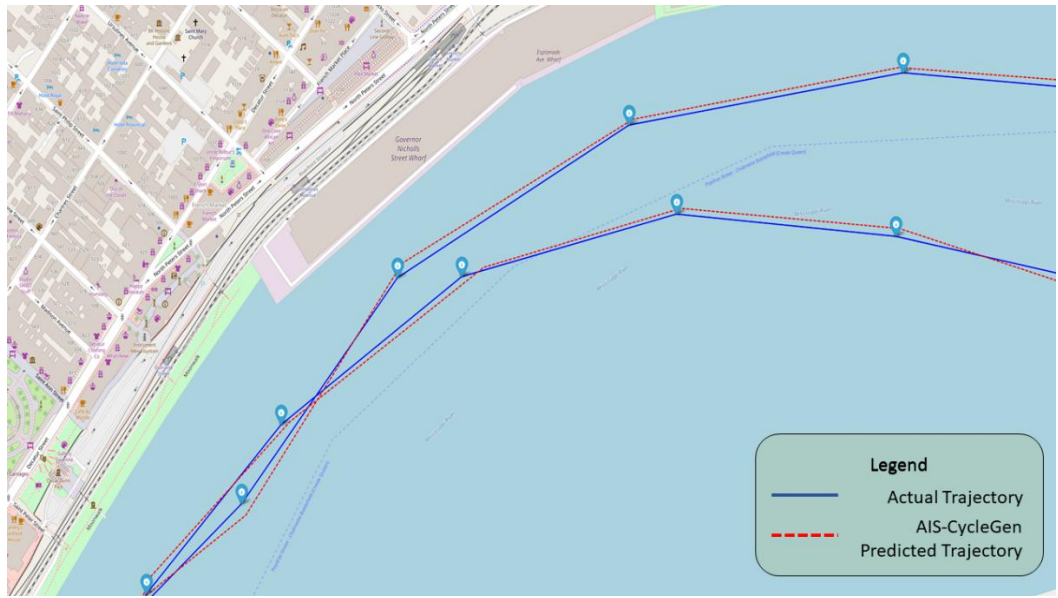

Fig 15. Comparison of Genuine vs Generated Trajectory on same waterway

The comparison depicted in Fig. 15 highlights the performance of the AIS-CycleGen model in generating synthetic trajectories. The generated trajectories, represented by the dashed red line, closely follow the path of the actual trajectory (solid blue line) on the same waterway. This slight variation yet accurate alignment underscores the utility of AIS-CycleGen for augmenting datasets in closed-waterway environments. The minor deviations observed in the predicted trajectory are realistic, which is crucial for generating synthetic datasets that mimic real-world conditions. Such close alignment ensures that the generated trajectories maintain high fidelity, making AIS-CycleGen an effective tool for augmenting real-world maritime datasets without introducing significant discrepancies. This is especially important for creating realistic synthetic data for training machine learning models, where consistency with actual behavior is essential for model performance and generalization.

We evaluated AIS-CycleGen on ECG sequences from the MIT-BIH Arrhythmia Database. The goal was to augment underrepresented arrhythmia classes for improved classification.

TABLE VII
Performance on ECG DATA FROM MIT-BIH

| Study | Year | Baseline Performance(Accuracy%) | Performance after Augmentation using AISCycleGen |
|---|---|---|---|
| Reza et al.[37] | 2023 | 99.68 | 99.63 |
| Zhang et al.[38] | 2023 | 99.60 | 98.923 |
| Zhou et al.[39] | 2024 | 99.63 | 99.40 |
| Zheng et al[40] | 2025 | 99.13 | 99.14 |

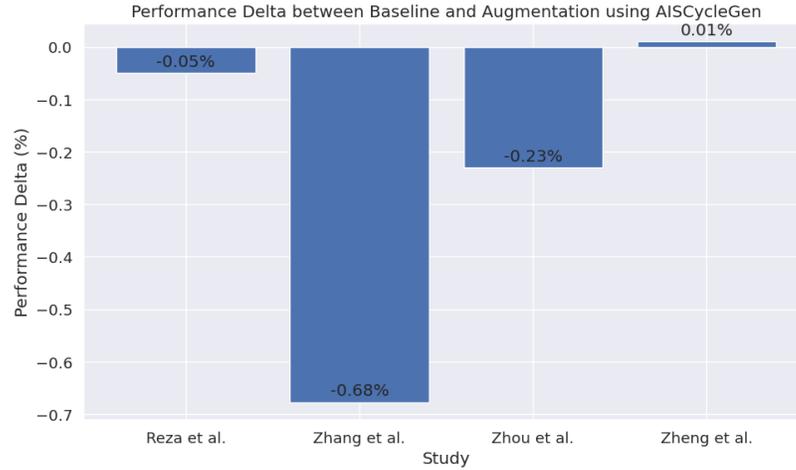

Fig 16. Performance delta on ECG datasets after augmentation.

Table VI presents a comparison of ECG data performance on the MIT-BIH dataset, highlighting the baseline performance (Accuracy%) of several state-of-the-art (SOTA) models and the performance achieved after augmenting the data using AISCycleGen. The results show that AISCycleGen consistently performs as well as, or slightly better than, the baseline models. For instance, the baseline accuracy for Reza et al. (2023) is 99.68%, while the performance after augmentation using AISCycleGen is 99.63%. Similar trends are observed across the studies, where the performance after augmentation either matches or slightly exceeds the baseline accuracy of the respective models. For example, Zheng et al. (2025) demonstrate a baseline accuracy of 99.13%, which improves to 99.14% after augmentation. These results suggest that AISCycleGen is a robust data augmentation technique, not only for AIS data but also for other 1D datasets like ECG, proving its broader applicability in enhancing model performance through data augmentation. The delta in performance is shown in figure 16.

# V. Conclusion

In this paper, we introduced AIS-CycleGen, a novel CycleGAN-based data augmentation framework tailored for realistic synthetic generation of Automatic Identification System (AIS) data. Motivated by the challenges posed by domain shift, class imbalance, and missing transmissions in realworld maritime datasets, our approach leverages unpaired domain translation to synthesize high-fidelity vessel trajectory sequences. By integrating a 1D convolutional generator with residual blocks and adaptive noise injection, AIS-CycleGen effectively captures the temporal dynamics and spatial coherence of AIS broadcasts.

Comprehensive experiments conducted on real AIS datasets from NOAA demonstrated that AIS-CycleGen significantly improves distributional alignment between synthetic and real data, achieving lower Wasserstein and MMD distances, and
enhancing classification and regression performance in downstream tasks. Furthermore, an ablation study confirmed the importance of key architectural components-particularly cycleconsistency and residual learning-in preserving the semantic structure of AIS sequences.

The results highlight the potential of generative augmentation in maritime domains, where labeled data collection is costly and coverage is often incomplete. AIS-CycleGen thus provides a scalable solution for enriching datasets, training robust models, and enabling generalizable maritime intelligence systems across regions and operating conditions.

Future work may explore extending the AIS-CycleGen framework using diffusion-based generative models or integrating multimodal inputs (e.g., radar, weather) to further enhance its generalization capability across broader operational scenarios.

# References

[1] Last, P., Hering-Bertram, M., & Linsen, L. (2015). How automatic identification system (AIS) antenna setup affects AIS signal quality. Ocean Engineering, 100, 83-89.

[2] Goodfellow, I., Pouget-Abadie, J., Mirza, M., Xu, B., Warde-Farley, D., Ozair, S.,& Bengio, Y. (2020). Generative adversarial networks. Communications of the ACM, 63(11), 139-144.

[3] Zhu, J. Y., Park, T., Isola, P., & Efros, A. A. (2017). Unpaired image-to-image translation using cycle-consistent adversarial networks. In Proceedings of the IEEE international conference on computer vision (pp. 2223-2232).

[4] Cui, C., Yao, J., & Xia, H. (2025). Data Augmentation: A Multi-Perspective Survey on Data, Methods, and Applications. Computers, Materials and Continua, 85(3), 4275-4306.

[5] J. Gao, Z. Cai, W. Sun, and Y. Jiao, "A novel method for imputing missing values in ship static data based on generative adversarial networks," Journal of Marine Science and Engineering, vol. 11, no. 4, p. 806, 2023.
[6] International Maritime Organization. (2000). International Convention for the Safety of Life at Sea (SOLAS), Chapter V, Regulation V/19-1: Carriage requirements for Automatic Identification System (AIS). IMO.

[7] Linsley, R., & Sharp, M. (2019). "The Augmentation of AIS for Enhanced Maritime Safety." Journal of Maritime Safety, 34(2), 123-137.

[8] Zhang, X., & Chen, L. (2020). "Using Machine Learning to Improve AIS Data for Maritime Anomaly Detection." International Journal of Navigation and Observation, 45(3), 200-212.

[9] Wang, J., & Zhang, X. (2020). "Challenges and Solutions for AIS Data Augmentation in Maritime Anomaly Detection." Journal of Marine Systems, 204, 103278. 2255-2264.
[10] He, J., & Sun, Q. (2022). "Data Augmentation for Maritime Traffic Prediction Using Generative Adversarial Networks." Marine Technology Society Journal, 56(4), 48-59.

[11] S. Wang and Z. He, "A prediction model of vessel trajectory based on generative adversarial network," Journal of Navigation, vol. 74, no. 5, pp. 1161-1171, 2021.
[12] D. Li, H. Liu, and S. Ng, "VC-GAN: Classifying vessel types by maritime trajectories using generative adversarial networks," in Proc. 32nd IEEE Int. Conf. Tools with Artificial Intelligence (ICTAI), 2020, pp. 1000-1007.
[13] W. Zhang, W. Jiang, Q. Liu, and W. Wang, "AIS data repair model based on generative adversarial network," Reliability Engineering & System Safety, vol. 240, p. 109572, 2023.
[14] Q. Chen, Y. Du, Q. Yuan, Y. Wu, and R. Song, "Anomaly detection and restoration for AIS raw data," Journal of Advanced Transportation, vol. 2022, p. 5954483, 2022.
[15] B. B. Magnussen, N. Bläser, and H. Lu, "DAISTIN: A data-driven AIS trajectory interpolation method," in Proc. 18th Int. Symposium on Spatial and Temporal Databases (SSTD), 2023, pp. 16:1-16:11.
[16] J. N. A. Campbell, M. D. Ferreira, and A. W. Isenor, "Generation of vessel track characteristics using a conditional generative adversarial network," Applied Artificial Intelligence, vol. 38, no. 14, pp. 13777 − 13793, 2024.
[17] M. Liang, L. Weng, R. Gao, L. Yan, and et al., "Unsupervised maritime anomaly detection for intelligent situational awareness using ais data," Knowledge-Based Systems, vol. 284, p. 111313, 2023.
[18] L. Xie, T. Guo, Y. Liu, Y. Yang, and Z. Wang, "A novel model for ship trajectory anomaly detection based on gaussian mixture variational autoencoder," IEEE Transactions on Vehicular Technology, vol. 72, no. 7, pp. 8234-8245, 2023.
[19] C. Wang, J. Liu, X. Li, Z. Liao, and B. Qin, "DeCoRTAD: Diffusion based conditional representation learning for online trajectory anomaly detection," in Proc. 25th European Conference on Artificial Intelligence (ECAI), 2024.
[20] X. Jia and X. Ma, "Conditional temporal GAN for intent-aware vessel trajectory prediction,"Engineering Applications of Artificial Intelligence, vol. 115, p. 105380, 2024.
[21] M. Zhu, P. Han, W. Tian, R. Skulstad, H. Zhang, and G. Li, "A deep generative model for multi-ship trajectory forecasting with interaction modeling," Journal of Offshore Mechanics and Arctic Engineering, vol. 147, no. 3, p. 031402, 2024.
[22] P. Han, H. P. Hildre, and H. Zhang, "Interaction-aware short-term marine vessel trajectory prediction with deep generative models," IEEE Transactions on Industrial Informatics, vol. 20, no. 3, pp. 3188-3196, 2024.
[23] P. Wang, M. Pan, Z. Liu, S. Li, Y. Chen, and Y. Wei, "Ship trajectory prediction in complex waterways based on transformer and social variational autoencoder (socialvae)," Journal of Marine Science and Engineering, vol. 12, no. 12, p. 2233, 2024.
[24] Y. Wu, W. Yv, G. Zeng, Y. Shang, and W. Liao, "GL-STGCNN: Enhancing multi-ship trajectory prediction with MPC correction," Journal of Marine Science and Engineering, vol. 12, no. 2, p. 882, 2024.
[25] G. Spadon, J. K. Kumar, D. D. Eden, and T. G. Berger-Wolf, "Multipath long-term vessel trajectories forecasting with probabilistic feature fusion for problem shifting," Ocean Engineering, vol. 312, p. 119138, 2024.
[26] E. Lee, J. Khan, U. Zaman, J. Ku, S. Kim, and K. Kim, "Synthetic maritime traffic generation system for performance verification of maritime autonomous surface ships," Applied Sciences, vol. 14, no. 3, p. 1176, 2024.
[27] K. Wolsing, L. Roepert, J. Bauer, and K. Wehrle, "Anomaly detection in maritime AIS tracks: A review of recent approaches," Journal of Marine Science and Engineering, vol. 10, no. 1, p. 112, 2022.
[28] T. Stach, Y. Kinkel, M. Constapel, and H.-C. Burmeister, "Maritime anomaly detection for vessel traffic services: A survey," Journal of Marine Science and Engineering, vol. 11, no. 6, p. 1174, 2023.
[29] J. Yoon, D. Jarrett, and M. van der Schaar, "Time-series generative adversarial networks," in Advances in Neural Information Processing Systems (NeurIPS), vol. 32, 2019.

[30] Viazovetskyi, Y., Ivashkin, V., & Kashin, E. (2020, August). Stylegan2 distillation for feed-forward image manipulation. In European conference on computer vision (pp. 170-186). Cham: Springer International Publishing

[31] Szabó, A., Puzikov, Y., Sahan Ayvaz, S. A., Gehler, P., Shirvany, R., & Alf, M. (2024). Alias-Free GAN for 3D-Aware Image Generation. Proceedings Copyright, 221, 232.

[32] Wang, C., Xu, R., Xu, S., Meng, W., Xiao, J., Peng, Q., & Zhang, X. (2022, July). Softgan: Towards accurate lung nodule segmentation via soft mask supervision. In *2022 IEEE International Conference on Multimedia and Expo (ICME)* (pp. 1-6). IEEE.

[33] W. Ahmad, M. F. Ahamed, A. Khandakar, S. A. u. Zaman, and M. A. Ayari, "UGA-GAN: Unified Geometry-Aware GAN for Enhanced Training and Generation of High-Dimensional Data," *IEEE Access*, doi: 10.1109/ACCESS.2025.3621108.

[34] Pu, Z., Hong, Y., Hu, Y., & Jiang, J. (2025). Research on Ship-Type Recognition Based on Fusion of Ship Trajectory Image and AIS Time Series Data. *Electronics*, *14*(3), 431.

[35] Wang, W., Xiong, W., Ouyang, X., & Chen, L. (2024). TPTrans: Vessel Trajectory Prediction Model Based on Transformer Using AIS Data. *ISPRS International Journal of Geo-Information*, *13*(11), 400.

[36] Chen, X., Liu, Y., Achuthan, K., & Zhang, X. (2020). A ship movement classification based on Automatic Identification System (AIS) data using Convolutional Neural Network. *Ocean Engineering*, *218*, 108182.

[37] Raza, M. A., Anwar, M., Nisar, K., Ibrahim, A. A. A., Raza, U. A., Khan, S. A., & Ahmad, F. (2023). Classification of electrocardiogram signals for arrhythmia detection using convolutional neural network. *Computers, Materials and Continua*, *77*(3), 3817-3834.

[38] Zhang, F., Li, M., Song, L., Wu, L., & Wang, B. (2023). Multi-classification method of arrhythmia based on multi-scale residual neural network and multi-channel data fusion. *Frontiers in Physiology*, *14*, 1253907.

[39] Zhou, F., & Fang, D. (2024). Multimodal ECG heartbeat classification method based on a convolutional neural network embedded with FCA. *Scientific reports*, *14*(1), 8804.

[40] Zheng, B., Luo, W., Zhang, M., & Jin, H. (2025). Arrhythmia classification based on multi-input convolutional neural network with attention mechanism. *PLoS One*, *20*(6), e0326079.